\documentclass{article}

    \PassOptionsToPackage{numbers, compress}{natbib}

     \usepackage[preprint]{neurips_2019}

\usepackage[utf8]{inputenc} 
\usepackage[T1]{fontenc}    
\usepackage{hyperref}       
\usepackage{url}            
\usepackage{booktabs}       
\usepackage{amsfonts}       
\usepackage{nicefrac}       
\usepackage{microtype}      
\usepackage{times}
\usepackage{epsfig}
\usepackage{graphicx}
\usepackage{amsmath}
\usepackage{amssymb}
\usepackage{enumitem}
\usepackage{subcaption}
\usepackage{caption}
\usepackage{multirow}
\usepackage{diagbox}
\usepackage[]{algorithm2e}
\usepackage{wrapfig}

\usepackage{floatrow}
\usepackage{xspace}
\newcommand{\coco}{MS-COCO\xspace} 

\def\onedot{\ifx\let@token.\else.\null\fi\xspace}
\def\eg{\emph{e.g}\onedot}

\newcommand{\plusminus}[1]{{\scriptstyle ~\pm~ #1}}

\title{One-Shot Instance Segmentation}

\author{Claudio Michaelis\\
\and
Ivan Ustyuzhaninov\\
\and
Matthias Bethge\\
\and
Alexander S. Ecker\\
\and
University of T\"ubingen\\
{\tt\small claudio.michaelis@uni-tuebingen.de}
}

\begin{document}

\maketitle

\begin{abstract}
   We tackle the problem of one-shot instance segmentation: Given an example image of a novel, previously unknown object category (the \emph{reference}), find and segment all objects of this category within a complex scene (the \emph{query image}).
   To address this challenging new task, we propose Siamese Mask R-CNN. It extends Mask R-CNN by a Siamese backbone encoding both reference image and scene, allowing it to target detection and segmentation towards the reference category.
   We demonstrate empirical results on \coco highlighting challenges of the one-shot setting: while transferring knowledge about instance segmentation to novel object categories works very well, targeting the detection network towards the reference category appears to be more difficult.
   Our work provides a first strong baseline for one-shot instance segmentation and will hopefully inspire further research into more powerful and flexible scene analysis algorithms. Code is available at: \url{https://github.com/bethgelab/siamese-mask-rcnn}
\end{abstract}

\begin{figure}[h]
    \centering
    \includegraphics[width=\linewidth]{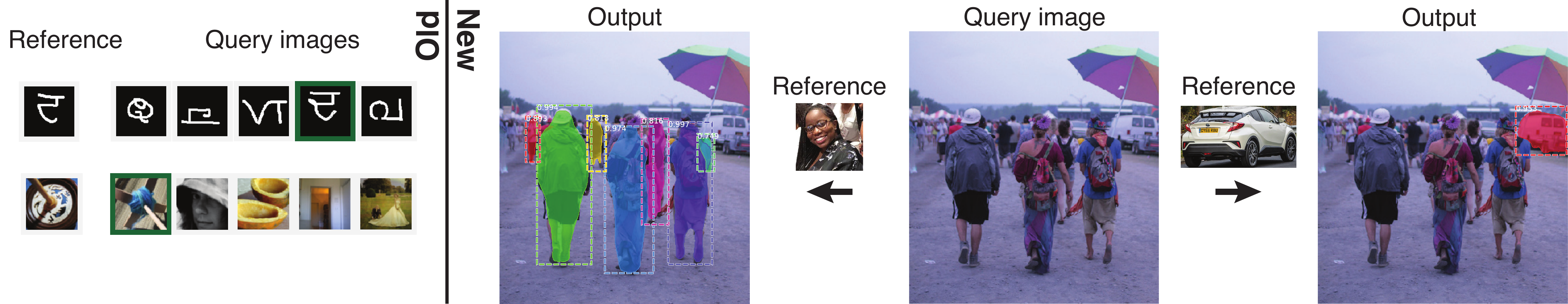}
    \caption{\textbf{Left:} Classical one-shot learning tasks are phrased as multi-class discrimination on datasets such as Omniglot and \emph{mini}Imagenet. \textbf{Right:} We propose one-shot instance segmentation on \coco.
    The bounding boxes and instance masks are outputs of our model.}
    \label{fig:teaser}
\end{figure}

\section{Introduction}
\label{sec:introduction}


Humans do not only excel at acquiring novel concepts from a small number of training examples (\emph{few-shot learning}), but can also readily point to such objects (\emph{object detection}) and draw their outlines (\emph{instance segmentation}). Conversely strong machine vision algorithms exist which can detect and segment a limited number of object categories in complex scenes \cite{Redmon2018, Lin2017b, He2017}. However in contrast to humans they are unable to incorporate new object concepts for which only a small number of training examples are provided. Enabling these object detection and segmentation systems to perform few-shot learning would be extremely useful for many real-world applications for which no large-scale annotated datasets like \coco~\cite{Lin2014} or OpenImages~\cite{Krasin2017} exist. Examples include autonomous agents such as household, service or manufacturing robots, or detecting objects in images collected in scientific settings (e.\,g.\ medical imaging or satellite images in geosciences).

Computer vision has made substantial progress in few-shot learning in the last years \cite{Lake2015, Snell2017, Finn2017, Rusu2018, Liu2019}. However, the field has focused on image classification in a discriminative setting, using datasets such as Omniglot~\cite{Lake2015} and MiniImagenet~\cite{Vinyals2016} (see Figure~\ref{fig:teaser}, left). As a consequence, these approaches are limited to rather simple object-centered images and cannot trivially handle object detection.

In this paper, we combine few-shot learning and instance segmentation in one task: We learn to detect and segment arbitrary objects in complex real-world scenes based on a single visual example (Figure~\ref{fig:teaser}, right). That is, we want our system to be able to find people and cars even though it has been provided with only one (or a few) labeled examples for each of those object categories. 

To evaluate the success of such a system, we formulate the task of one-shot instance segmentation: Given a scene image and a previously unknown object category defined by a single reference instance, generate a bounding box and a segmentation mask for every instance of that category in the image. 
This task can be seen as an example-based version of the typical instance segmentation setup and is closely related to the everyday problem of visual search which has been studied extensively in human perception \cite{Sanders1996, Wolfe2011}.

We show that a new model, \emph{Siamese Mask R-CNN}, which incorporates ideas from metric learning (Siamese networks~\cite{Koch2015}) into Mask R-CNN~\cite{He2017}, a state-of-the-art object detection and segmentation system (Figure~\ref{fig:siamese_mask_rcnn}), can learn this task and acquire a similarity metric that allows it to generalize to previously unknown object categories.

Our main contributions are:
\begin{itemize}[nosep]
    \item We introduce one-shot instance segmentation, a novel one-shot task, requiring object detection and instance segmentation based on a single visual example.
    \item We present \emph{Siamese Mask R-CNN}, a system capable of performing one-shot instance segmentation.
    \item We establish an evaluation protocol for the task and evaluate our model on \coco.
    \item We show that, for our model, targeting the detection towards the reference category is the main challenge, while segmenting the correctly identified objects works well.
\end{itemize}

\section{Background}

\paragraph{Object detection and instance segmentation.}

In computer vision, object detection is the task of localizing and classifying individual objects in a scene \cite{Everingham2010}. It is usually formalized as: Given an image (\emph{query image}), localize all objects from a fixed set of categories and draw a bounding box around each of them. Current state-of-the-art models use a convolutional neural network (the \emph{backbone}) to extract features from the query image and subsequently classify the detected objects into one of the $n$ categories (or background). Most models either directly use the backbone features to predict object locations and categories (\emph{single stage}) \cite{Liu2016, Redmon2016, Redmon2017, Redmon2018, Lin2017b} or first generate a set of class-agnostic object proposals which are subsequently classified (\emph{two stage}) \cite{Girshick2014, Girshick2015, Ren2015, He2017}.

Segmentation tasks require labeling all pixels belonging to a certain semantic category (\emph{semantic segmentation}) or object instance (\emph{instance segmentation}). While both tasks seem closely related, they in fact require quite different approaches: Semantic segmentation models perform pixel-wise classification and are usually implemented using fully convolutional architectures \cite{Long2015, Noh2015, Ronneberger2015, Zhao2017, Chen2018b}.
In contrast, instance segmentation is more closely related to object detection, as it requires identifying individual object instances \cite{Hariharan2014, Dai2015, Pinheiro2015, Ren2017, He2017}. It therefore inherits the difficulties of object detection, which make it a significantly harder task than semantic segmentation. Consequently, the current state-of-the-art instance segmentation model (\emph{Mask R-CNN}) \cite{He2017} is an extension of a successful object detection model (\emph{Faster R-CNN}) \cite{Ren2015}.

\paragraph{Few-shot learning}

The goal of few-shot learning is to find models which can generalize to novel categories from few labeled examples \cite{FeiFei2006, Lake2015}. This capability is usually evaluated through a number of \emph{episodes}. Each episode consists of a few examples from novel categories (the \emph{support set}) and a small test set of images from the same categories (the \emph{query set}). When the support set contains $k$ examples from $n$ categories, the problem is usually referred to as an \emph{n-way, k-shot} learning problem. In the extreme case when only a single example per category is given, this is referred to as one-shot learning.

There are two main approaches to solve this task: either train a model to learn a metric, based on which examples from novel categories can be classified (\emph{metric learning})~\cite{Koch2015, Vinyals2016, Snell2017, Wang2018} or to learn a good learning strategy which can be applied in each episode (\emph{meta learning})~\cite{Finn2017, Li2017a, Munkhdalai2017, Munkhdalai2018, Sung2018, Ren2018, Sun2019, Rusu2018}. To train these models, the categories in a dataset are usually split into \emph{training} categories used to train the models and \emph{test} categories used during the evaluation procedure. Therefore, the few-shot model will be trained and tested on different categories, forcing it to generalize to novel categories.

\section{One-shot object detection and instance segmentation on \coco }
\label{sec:task}

The goal of one-shot object detection and instance segmentation is to develop models that can localize and segment objects from arbitrary categories when provided with a single visual example from that category. To this end, we 1) replace the widely used category-based object detection task by an example-based task setup and 2) split the available object categories into a training set and a non-overlapping test set, which is used to evaluate generalization to unknown categories. We use the popular \coco dataset, which consists of a large variety of complex scenes with multiple objects from abroad range of categories and often challenging conditions like clutter.

\paragraph{Task setup: example-based instance segmentation.}

We define one-shot detection and segmentation as follows: Given a \emph{reference image} showing a close-up example of a novel object category, find and segment all instances of objects belonging to this category in a separate \emph{query image}, which shows an entire visual scene containing many objects (Figure~\ref{fig:teaser}, right). The main difference between this task and the usual object detection setup is the change from a category-based to an example-based setup. Instead of requiring to localize objects from a number of fixed categories, the example-based task requires to detect objects from a single category, which is defined through a reference image. The reference image shows a single object instance of the category that is to be detected, cropped to its bounding box (see Figure~\ref{fig:teaser} for two examples). It is provided without mask annotations.

\paragraph{Split of categories for training and testing.}

To be able to evaluate performance on novel categories, we split the 80 object categories in \coco into 60 \emph{training} and 20 \emph{test} categories.
Following earlier work on Pascal VOC \cite{Shaban2017}, we generate four such training/test splits by including every fourth category into the test split, starting with the first, second, third or fourth category, respectively (see Table~\ref{table:dataset_splits} in the Appendix).

Because we use complex scenes which can contain objects from many categories, it is not feasible to ensure that the training images contain no instances of held-out categories. However, we do not provide any annotations for these categories during training and never use them as references. In other words, the model will see objects from the test categories during training, but is never provided with any information about them. This setup differs from the typical few-shot learning setup, in which the model never encounters any instance of the novel objects during training. However, in addition to being the only feasible solution, we consider this setup quite realistic for an autonomous agent, which may encounter unlabeled objects multiple times before they become relevant and label information is provided. Think of a household robot seeing, but not recognizing, a certain type of toy in various parts of the apartment multiple times before you instruct it to go pick it up for you.

\paragraph{Evaluation procedure.}

We propose to evaluate task performance using the following procedure:
\begin{enumerate}[nosep]
    \item Choose an image from the test set
    \item Draw a random reference image for each of the (novel) test categories present in the image
    \item Predict bounding boxes for each reference image separately
    \item Assign the computed predictions to the category of the corresponding reference image
    \item Repeat this process for all images in the test set
    \item Compute mAP50 \cite{Everingham2010} using the standard tools from object detection \cite{Coco2018}~\footnote{We chose to use mAP50 (mAP @ 50\% Bounding Box IoU~\cite{Everingham2010}) instead of the COCO metric mAP (mean of mAP @ 50, 55, ..., 95\% Bounding Box IoU~\cite{Lin2014}), because we think it more directly reflects the result we are primarily interested in: whether our model can find novel objects based on a single reference image. For results using the \coco metric see Appendix Section~\ref{appendix:additional_results}}
\end{enumerate}

The same steps as above apply in the case of instance segmentation, with the difference that a segmentation mask instead of a bounding box is required for each predicted object.

Our evaluation procedure is simplified somewhat, because we ensure that the reference categories are actually present in each image used for evaluation. For a real-world application of such a system, this restriction would have to be removed. However, we found the task to be very challenging already with this simplification, so we believe it is justified for the time being.

\paragraph{Connection to few-shot learning and object detection.}

Our evaluation procedure lends from other few-shot setups that typically evaluate in episodes. Each episode consists of a support set (the training examples for the novel categories) and a query set (the images to be classified). In our case, an episode consists of the detection of objects of one novel category in one image. In this case, the support set is the set of examples from the category to be detected (the \emph{references}) while the query set is a single image (the \emph{query image}).
Compared to object detection, the classifier is turned into a binary verification conditioned on the reference image(s). Compared to the typical few-shot learning setup, there are two key differences: First, as only one category is given, the task is not a discrimination task between the given categories, but a verification task between the given category and all other object categories. Second the query image may not only contain objects from the novel category given by the reference, but also other objects from known and unknown categories.

\paragraph{Connection to other related tasks.} 

Our setup differs from a number of related paradigms. In contrast to recent work on few-shot object detection \cite{Dong2018, Chen2018, kang2018few, Schwartz2018a}, we formulate our task as an example-based search task rather than learning an object detector from a small labeled dataset. This allows us to directly apply our model on novel categories without any retraining. We also extend all of these approaches by additionally asking the system to output segmentation masks for each instance and focus on the challenging \coco dataset. Similarly our task shares similarities with zero-shot object detection~\cite{Bansal2018, Rahman2018, Demirel2018, zhu2019}, but with the crucial difference that in zero-shot detection the reference category is defined by a textual description instead of an image.

A range of one-shot segmentation tasks exist, including one-shot semantic segmentation~\cite{Shaban2017,Rakelly2018, Dong2018a, Michaelis2018}, texture segmentation~\cite{Ustyuzhaninov2018}, medical image segmentation~\cite{Zhao2019} and recent work on co-segmentation~\cite{Li2018}\footnote{Most co-segmentation work (e.g.~\cite{Rother2006, Faktor2013}) uses the same object categories during training and test time and therefore does not operate in the few-shot setting}.
The key difference is that the models developed for these tasks output pixel-level semantic classifications rather than instance-level masks and, thus, cannot distinguish individual object instances. In co-segmentation very recent work \cite{Hsu2019} explores instance co-segmentation, but not in a few-shot setting.
Two studies segment instances in a few-shot setting, but with different task setups: (1) in one-shot video segmentation~\cite{Caelles2017, Caelles2019}, object instances are tracked across a video sequence; (2) in one-shot instance segmentation of homogeneous object clusters~\cite{wu2018annotation} a model is proposed which segments, e.\,g., a pile of bricks into the individual instances based on a video pan of one of the bricks. Both of these setups are closer to particular object retrieval \cite{Tolias2016, Salvador2016, Gordo2016}, as they localize instances of a particular object rather than instances of the same object category, as is the focus of our work.

\section{Siamese Mask R-CNN}
\label{sec:model}

The key idea of one-shot instance segmentation is to detect and segment object instances based on a single visual example of some object category. Thus, our system has to deal with arbitrary, potentially previously unknown object categories which are defined only through a single reference image, rather than with a fixed set of categories for which extensive labeled data was provided during training. To solve this problem, we take a metric-learning approach: we learn a similarity metric between the reference and image regions in the scene. Based on this similarity metric, we then generate object proposals and classify them into matches and non-matches. The key advantage of this approach is that it can be directly applied to objects of novel categories without the need to retrain or fine-tune the learned model.

\begin{figure}[t]
    \centering
    \includegraphics[width=\linewidth]{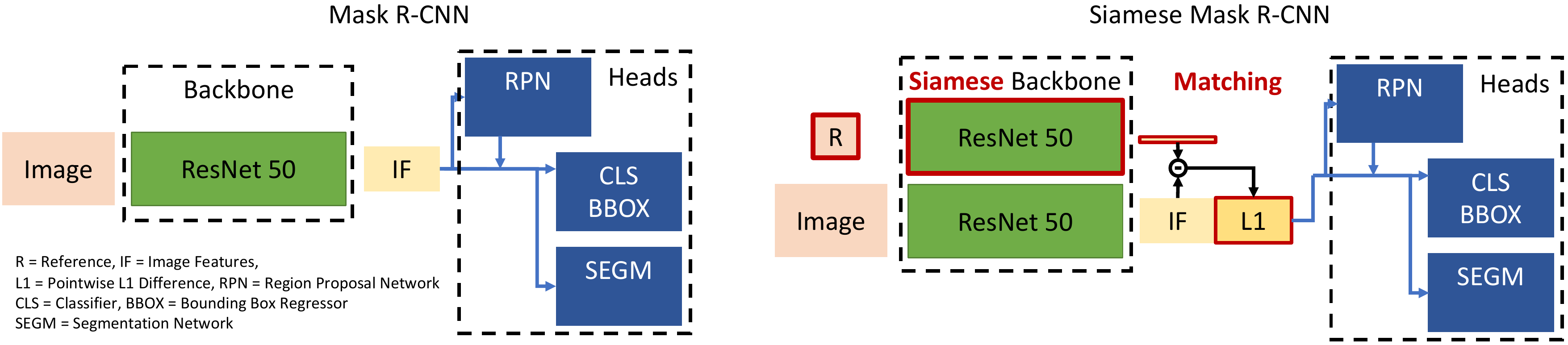}
    \caption{Comparison of Mask R-CNN and Siamese Mask R-CNN. The main differences (marked in red) of our model are (1) the Siamese backbone which jointly encodes the image and reference, and (2) the matching of those embeddings to target the region proposal and classification heads towards the reference category.}
    \label{fig:siamese_mask_rcnn}
\end{figure}

To compute the similarity metric we use Siamese networks, a classic metric learning approach ~\cite{Bromley1993, Chopra2005, Koch2015}. We combine this form of similarity judgment with the domain knowledge built into current state-of-the-art object detection and instance segmentation systems by integrating it into Mask R-CNN~\cite{He2017}. In the following paragraphs we provide a quick recap of Mask R-CNN before describing the changes we made to integrate the Siamese approach and how we compute the similarity metric. We build our implementation upon the Matterport Mask R-CNN library~\cite{Abdulla2017}. The details can be found in Appendix~\ref{appendix:implementation} and in our code\footnote{\url{https://github.com/bethgelab/siamese-mask-rcnn}}.

\paragraph{Mask R-CNN.}

Mask R-CNN is a two-stage object detector that consists of a backbone feature extractor and multiple heads operating on these features (see Figure~\ref{fig:siamese_mask_rcnn}).
The heads consist of two stages. First, the region proposal network (RPN) is applied convolutionally across the image to predict possible object locations in the scene. The most promising region proposals are then cropped from the backbone feature maps and used as inputs for the bounding box classification (CLS) and regression (BBOX) head as well as the instance masking head (MASK).

\paragraph{Siamese network backbone.}

To integrate the reference information into Mask R-CNN, the same backbone (ResNet50 \cite{He2016} with Feature Pyramid Networks (FPN) \cite{Lin2017}) is used with shared weights to extract features from both the reference and the scene.

\paragraph{Feature matching.}

\begin{wrapfigure}[7]{r}{0.5\textwidth}
    \vspace{-3.5mm}
    \includegraphics[width=\linewidth]{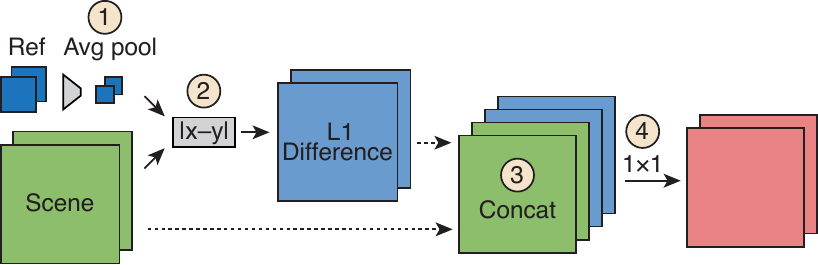}
    \caption{Sketch of the matching procedure. 
    }
    \label{fig:matching}
\end{wrapfigure}

To obtain a measure of similarity between the reference and different regions of the query image, we treat each (x,y) location of the encoded features of the query image as an embedding vector and compare it to the embedding of the reference image. This procedure can be viewed as a non-linear template matching in the embedding space instead of the pixel space. The matching procedure works as shown in Figure~\ref{fig:matching}:
\begin{enumerate}[nosep]
    \item Average pool the features of the reference image to an embedding vector. In the few-shot case (more than one reference) compute the average of the reference features as in prototypical networks \cite{Snell2017}.
    \item Compute the absolute difference between the reference embedding and that of the scene at each (x,y) position.
    \item Concatenate this difference to the scene representation.
    \item Reduce the number of features with a $1 \times 1$ convolution.
\end{enumerate}

The resulting features are then used as a drop-in replacement for the original Mask R-CNN features \footnote{As we use  a backbone with feature pyramid networks (FPN) we get features at multiple resolutions. We therefore simply apply the described matching procedure at each resolution independently.}. The key difference is that they do not only encode the content of the scene image, but also its similarity to the reference image, which forms the basis for the subsequent heads to generate object proposals, classify matches vs. non-matches and generate instance masks.

\paragraph{Head architecture}

Because the computed features can be used as a drop-in replacement for the original features, we can use the same region proposal network and ROI pooling operations as Mask R-CNN. We can also use the same classification and bounding box regression head as Mask R-CNN, but change the classification from an 80-way category discrimination to a binary match/non-match discrimination and generate only a single, class-agnostic set of bounding box coordinates. Similarly, for the mask branch we predict only a single instance mask instead of one per potential category.

\section{Experiments}
\label{sec:experiments}

We train Siamese Mask R-CNN jointly on object detection and instance segmentation in the example-based setting using the training set of \coco.
We train one model on each of the four category splits defined in Section~\ref{sec:task} and evaluate the trained models on both known (train) and unknown (test) categories using the \coco validation set. In the following paragraphs, we highlight the most important changes between our training and evaluation protocol and that of Mask R-CNN. The full training and evaluation details are given in Appendix~\ref{appendix:training} and~\ref{appendix:evaluation}.

\paragraph{Training.}
\label{subsec:training}

We first pre-train the ResNet backbone on a reduced subset of ImageNet, which contains only images from the 687 ImageNet categories that have no correspondence in \coco. We do this to avoid using any label information about the test categories during pre-training.

We then proceed by training episodically. For each image in a minibatch, we pick a random reference category among the training categories present in the image. We then crop a random instance of this category out of another random image in the training set. We keep only the annotations of this category; all other objects are treated as background.

\paragraph{Evaluation.}

We evaluate our model using the procedure described in Section~\ref{sec:task}. Each category split is evaluated separately. The final score is the mean of the scores from all four splits. This evaluation procedure is stochastic due to the random selection of references. We thus repeat the evaluation five times and report the average and 95\% confidence intervals.

\paragraph{Baseline: random boxes.}
As a simple sanity check, we evaluate the performance of a model predicting random bounding boxes and segmentation masks. To do so, we take ground-truth bounding boxes and segmentation masks for the category of the reference image, and randomly shift the boxes around the image (assigning a random confidence value for each box between 0.8 and 1). We keep the ground-truth segmentation masks intact in the shifted boxes. This procedure allows us to get random predictions while keeping certain statistics of the ground-truth annotations (\eg number of boxes per image, their sizes, etc.).

\begin{figure*}[t]
    \begin{center}
    \includegraphics[width=\linewidth]{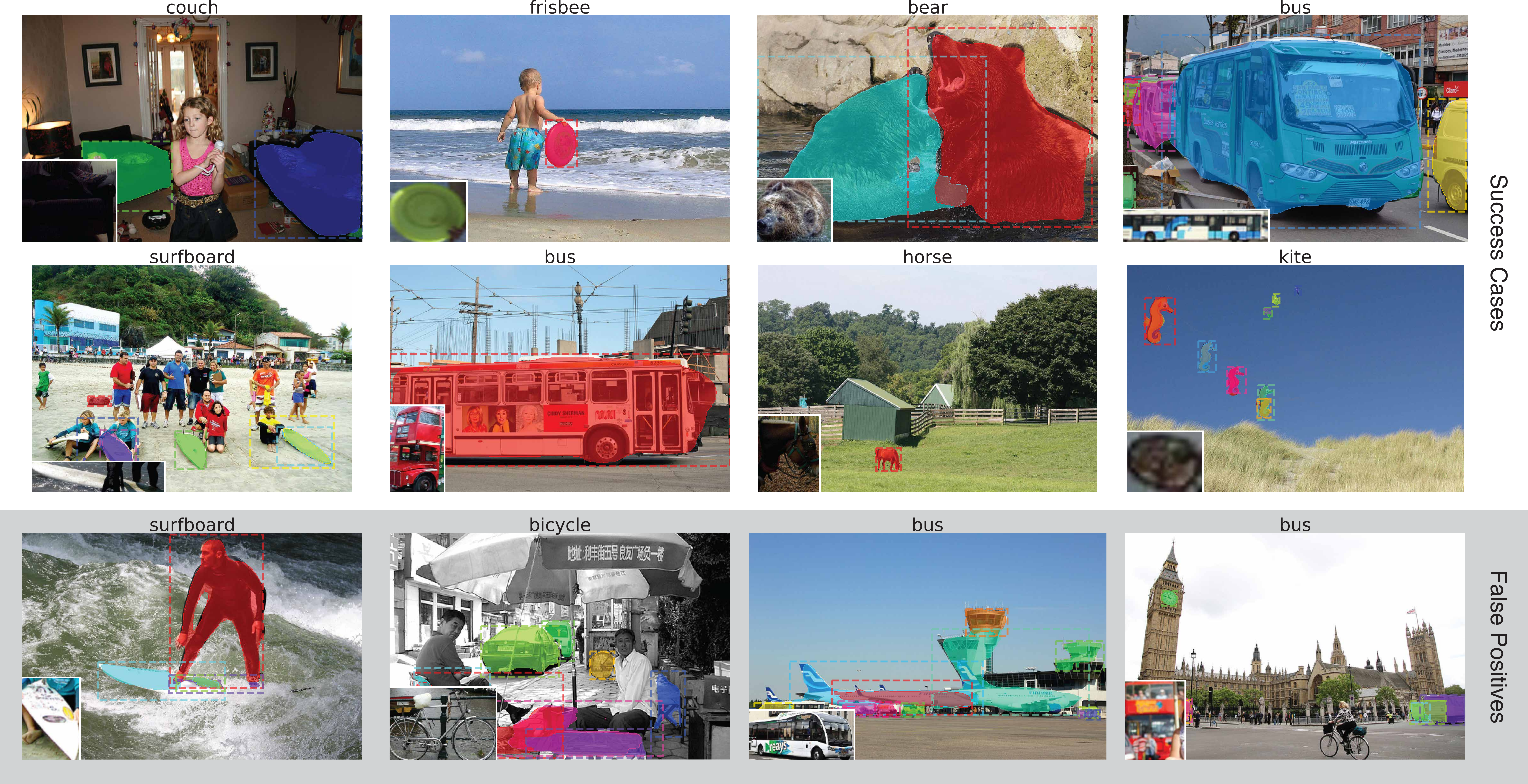}
    \end{center}
    \vspace{-0.2cm}
    \caption{Examples of Siamese Mask R-CNN operating in the one-shot setting, i.e. segmenting novel objects which are not known from training (split $S_2$). The only information our model has about these categories is one reference image (shown in the lower-left corner of each example; the categories in the titles are just for the reader). The top two rows show success cases while the last row displays some results with a lot of false positives. Best viewed with zoom and color.}
\label{fig:visual_examples}
\end{figure*}

\section{Results}
\label{sec:results}

\paragraph{Example-based detection and segmentation.}

We begin by applying the trained Siamese Mask R-CNN model to detect objects from the categories used for training. In this setting, all of the training examples are used to learn the metric, but the detection is based only on the similarity to one (or five) instance(s) from the reference category. IWith one reference, we achieve 37.6\% and 34.9\% mAP50 for object detection and instance segmentation, respectively. With five references, we achieve 41.3\% and 38.4\%, respectively (Table ~\ref{table:results_one_shot}). We also report the 95\% confidence interval estimated from five evaluation runs to quantify the variability introduced by to the random selection of reference images. The variation is below $0.2$ percentage points in all cases, which suggests that evaluating five times is sufficient to handle the variability. We observe some additional variation between the splits, which seems to stem mostly from the over-representation of the person category (see Appendix Table~\ref{table:results_by_split} for results of each split).

\paragraph{One-shot instance segmentation.}



Next, we report the results of evaluating Siamese Mask R-CNN on novel categories not used for training, showcasingits ability to generalize to the 20 held-out categories that have not been annotated during training. With one reference (one-shot), the average detection mAP50 score for the test splits is 16.3\%, while the segmentation performance is 14.5\% (Table~\ref{table:results_one_shot}).
While these values are significantly lower than those for the training categories, they still present a strong baseline and are far from chance (1.2\%/0.5\% for detection/segmentation) despite the difficulty of the one-shot setting. When using five references (five-shot), the performance improves to 18.5\% and 16.7\%, respectively. Taken together, these results suggest that the metric our model has learned allows some generalization outside of the training categories, but a substantial degree of overfitting on the those categories remains.

\begin{table}[t]
\begin{center}
\begin{small}
\begin{tabular}{lccccccccc}
& \hspace{5pt} & \multicolumn{2}{c}{Categories used in training} & \hspace{5pt} & \multicolumn{2}{c}{Novel categories} & \hspace{5pt} & Random \\
& & 1-shot & 5-shot & & 1-shot & 5-shot & &\\
\hline\hline
Object detection & & $37.6 \plusminus{0.2}$ & $41.3 \plusminus{0.1}$ & & $16.3 \plusminus{0.1}$ & $18.5 \plusminus{0.1}$ & & $1.2 \plusminus{0.1}$ \\
Instance segmentation & & $34.9 \plusminus{0.1}$ & $38.4 \plusminus{0.1}$ & & $14.5 \plusminus{0.1}$ & $16.8 \plusminus{0.1}$ & & $0.5 \plusminus{0.1}$
\end{tabular}
\end{small}
\end{center}
\caption{Results on \coco (in \% mAP50 with 95\% confidence intervals). Three settings are reported: Evaluating on training (train), novel (test) categories and randomly drawn boxes (random). We run our models with one or five references per category and image (shots).}
\label{table:results_one_shot}
\end{table}

\paragraph{Qualitative analysis.}

\begin{wrapfigure}[15]{r}{0.25\textwidth}
    \centering
    \includegraphics[width=\linewidth]{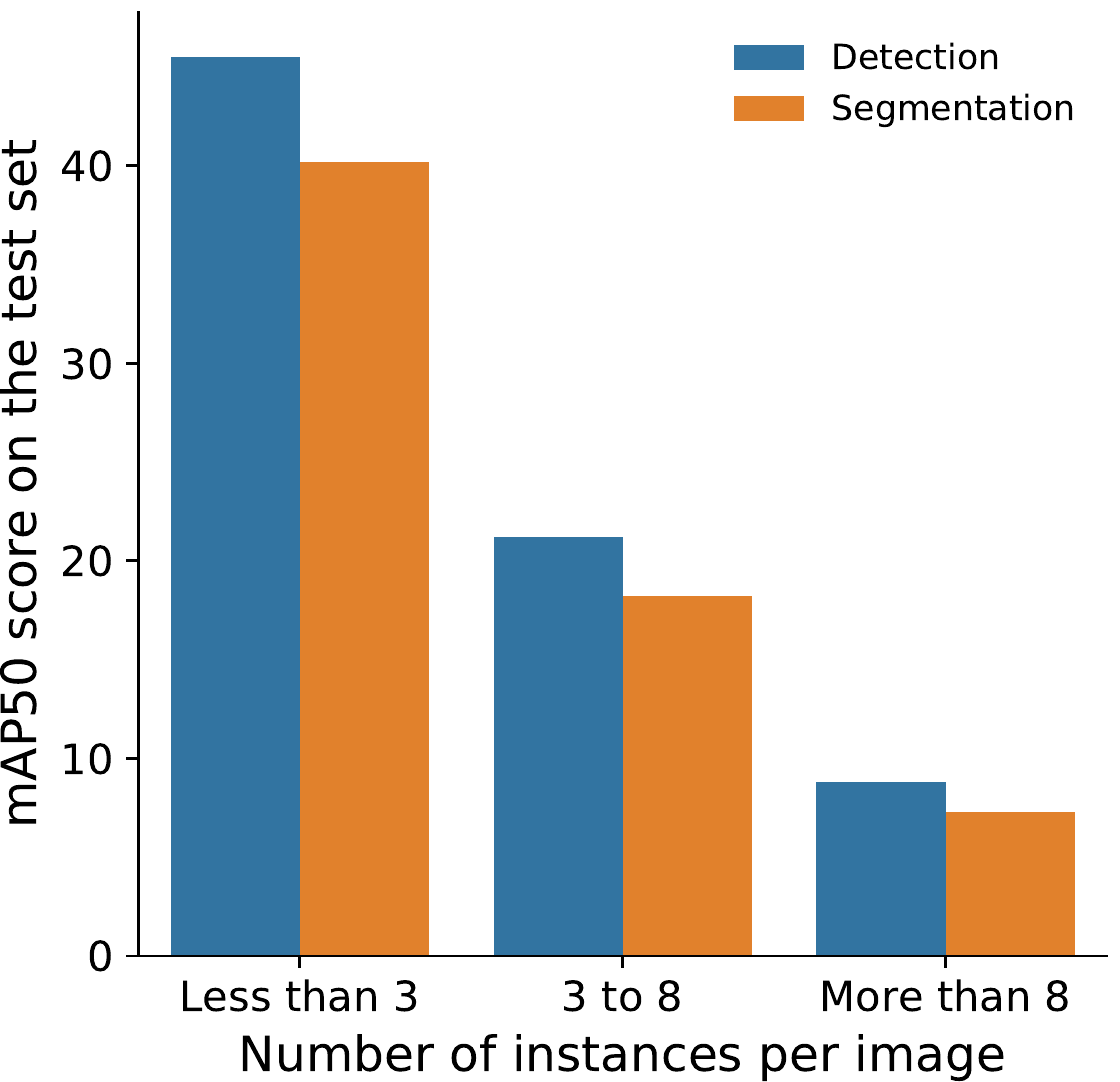}
    \vspace{-0.5cm}
    \caption{Results on split $S_2$ (in \% mAP50) separated by the number of instances per image.}
    \label{fig:clutter-results}
\end{wrapfigure}

The first two rows of Figure~\ref{fig:visual_examples} show some examples of successful detection and segmentation of objects from novel categories. These examples allow us to get a feeling for the difficulty of the task: the reference inputs are quite different from the instances in the query image, often showing different perspectives, usually very different instances of the category and sometimes only parts of the reference object. Also note that the ground truth segmentation mask is not used to pre-segment the reference.

To generate bounding boxes and segmentation masks, the model can thus use only its general knowledge about objects. It has to rely on the metric learned on the categories annotated during training to decide whether the reference and the query instances belong to the same category. For instance, the bus and the horse in the second row of Figure~\ref{fig:visual_examples} are incomplete and the network has never been provided with ground truth bounding boxes or instance masks for either horses or buses. Nevertheless, it still finds the correct object in the query image and segments the entire object.

We also show examples of failure cases in the last row of Figure~\ref{fig:visual_examples}. The picture that emerges from both successful and failure cases is that the network produces overall good bounding boxes and segmentation masks, but often fails at targeting them towards the correct category. We elaborate more on the challenges of the task in the following paragraphs.

\paragraph{False positives when evaluating on novel categories.}

There is a marked drop in model performance between evaluating on the categories used during training and the novel categories, suggesting some degree of overfitting to the training categories. If this is indeed the case, we would expect false positives to be biased towards these categories and, in particular, towards those categories that are most frequent in the training set. Qualitatively, this bias seems indeed to exist (Figure~\ref{fig:visual_examples}). We verified this assumption quantitatively by computing a confusion matrix between categories (Appendix Figure~\ref{fig:confusion_matrix}). The confusion matrix shows that objects from the training categories are often falsely detected when searching for objects of the novel categories. Among the most commonly falsely detected categories are people, cars, airplanes and clocks which are overrepresented in the dataset.

\paragraph{Effect of image clutter.}

Previous work on synthetic data found that cluttered scenes are especially challenging in example based one-shot tasks~\cite{Michaelis2018}. This effect is also present in the current context. Both detection and segmentation scores are substantially higher for images with a small number of total instances (Figure~\ref{fig:clutter-results}), underscoring the importance of extending the model to robustly process cluttered scenes.

\section{Related work}
\label{sec:related_work}

As outlined in section~\ref{sec:task}, our approach lies at the intersection of few-shot/metric learning, object detection/visual search, and instance segmentation. Each of these aspects has been investigated extensively. The novelty of our approach is the combination of all these aspects. A number of very recent and, to a large extent concurrent, works have started addressing few-shot detection. We review the most closely related work below. We are not aware of any previous work on category-based few-shot instance segmentation.

Dong et al.~\cite{Dong2018} train a semi-supervised few-shot detector on the 20 categories of Pascal VOC using roughly 80 annotated images, supplemented by a large set of unlabeled images. They train a set of models, each of which generates training labels for the other models by using high-confidence detections in the unlabeled images.
The low-shot transfer detector (LSTD)~\cite{Chen2018} fine-tunes an object detector on a transfer task with new categories using two novel regularization terms: one for background depression and one for knowledge transfer from the source domain to the target domain.
Kang et. al. \cite{kang2018few} extend a single-stage object detector -- YOLOv2 \cite{Redmon2017} -- by a category-specific feature reweighting that is predicted by a meta model, allowing them to incorporate novel classes with few examples.
Schwartz et. al. \cite{Schwartz2018a} replace the classification branch of Faster R-CNN with a metric learning module, which evaluates the similarity of each predicted box to a set of prototypes generated from the few provided examples. 
Very recent concurrent work~\cite{Zhang2019} evaluates the same task as we do for object detection on Pascal VOC using Faster R-CNN, although they employ separate feature fusions in the RPN and classifier head instead of the unified matching we employ.
Recent works on zero-shot detection~\cite{Bansal2018, Rahman2018, Demirel2018, zhu2019} use a similar approach to ours to target the detection towards a novel category, except that they learn a joint embedding for the query image and a textual description (instead of a visual description) of this novel category.

\section{Discussion}
\label{sec:discussion}

We introduced the task of \emph{one-shot instance segmentation} which requires models to generalize to object categories that have not been labeled during training in the challenging setting of instance segmentation. To address this task we proposed \emph{Siamese Mask R-CNN}, a model combining a state-of-the-art instance segmentation model (Mask R-CNN) with a metric learning approach (Siamese networks). This model can detect and segment objects from novel categories based on a single reference image. While our approach is not as successful on novel categories as on those used for training, it performs far above chance, showcasing it's ability to generalize to categories outside of the training set. Generally, it is expected from any reasonable learning system that it should perform better on object categories for which it has been trained with thousands of examples than for those encountered in a few-shot setting. Considering the difficulty of this problem, the performance of our model should provide a strong baseline and we hope that our work provides a first step towards visual search algorithms with human like flexibility. 

\clearpage

\section*{Acknowledgements}

We want to thank Mengye Ren, Jake Snell, James Lucas, Marc Law, Richard Zemel and Eshed Ohn-Bar for helpful discussion. 
This work was supported by the Deutsche Forschungsgemeinschaft (DFG, German Research Foundation) via grant EC 479/1-1 to A.S.E and the Collaborative Research Center (Projektnummer 276693517 -- SFB 1233: Robust Vision), by the German Federal Ministry of Education and Research through the Tübingen AI Center (FKZ 01IS18039A), by the International Max Planck Research School for Intelligent Systems (C.M. and I.U.), and by the Advanced Research Projects Activity (IARPA) via Department of Interior/Interior Business Center (DoI/IBC) contract number D16PC00003. The U.S. Government is authorized to reproduce and distribute reprints for Governmental purposes notwithstanding any copyright annotation thereon. Disclaimer: The views and conclusions contained herein are those of the authors and should not be interpreted as necessarily representing the official policies or endorsements, either expressed or implied, of IARPA, DoI/IBC, or the U.S. Government.

{\small

}

\clearpage

\section*{Changes to previous version}
Compared to the previous version (submitted to arxiv on 28 Nov 2018) this version additionally includes: 
\begin{itemize}[nosep]
    \item a different evaluation procedure evaluating each split 5-times and reporting the mean and 95\% confidence interval.
    \item five-shot results using a prototypical approach to accomodate multiple reference images.
    \item a background section introducing information and notation of object detection and few-shot learning tasks.
    \item discussion of concurrent work which was published on arxiv since the publication of the previous version \cite{kang2018few, Zhang2019}.
    \item detailed description of the training and evaluation process in the Appendix.
    \item results for all metrics evaluated on the \coco leaderboard to the Appendix.
\end{itemize}
Additionally to adding content we reworked large parts of the text to clarify the task setup the way we present related tasks and the corresponding solutions. We also update some of the figures, mainly combining the two figures for qualitative analysis into one figure which includes good and bad examples, adding a comparison with traditional few-shot learning tasks to the introduction figure and making the color coding in the model figure easier to understand.

\section*{Appendix}
\setcounter{section}{0}
\setcounter{figure}{0}
\setcounter{table}{0}
\renewcommand*{\theHsection}{chX.\the\value{section}}
\renewcommand*{\thesection}{A\the\value{section}}
\renewcommand*{\thefigure}{A\the\value{figure}}
\renewcommand*{\thetable}{A\the\value{table}}

\section{Training and testing categories}
\label{appendix:splits}

This section contains the description of the category splits from Section~\ref{sec:task} from the main paper as well as a table of those categories.

\subsection{Splits $S_1$-$S_4$}
To be able to evaluate performance on novel categories we hold out some categories during training. We split the 80 object categories in \coco into 60 \emph{training} and 20 \emph{test} categories.
Following earlier work on Pascal VOC \cite{Shaban2017}, we generate four such training/test splits by including every fourth category into the test split starting with the first, second, third or fourth category, respectively. These splits are shown in Table~\ref{table:dataset_splits} below.

\begin{table}[h]
\begin{tabular}{rl|rl|rl|rl}
\multicolumn{2}{c|}{$S_1$} & \multicolumn{2}{c|}{$S_2$} & \multicolumn{2}{c|}{$S_3$} & \multicolumn{2}{c}{$S_4$} \\ \hline
1       & Person            & 2      & Bicycle           & 3      & Car               & 4      & Motorcycle        \\
5       & Airplane          & 6      & Bus               & 7      & Train             & 8      & Truck             \\
9       & Boat              & 10     & Traffic light     & 11     & Fire Hydrant      & 12     & Stop sign         \\
13      & Parking meter     & 14     & Bench             & 15     & Bird              & 16     & Cat               \\
17      & Dog               & 18     & Horse             & 19     & Sheep             & 20     & Cow               \\
21      & Elephant          & 22     & Bear              & 23     & Zebra             & 24     & Giraffe           \\
25      & Backpack          & 26     & Umbrella          & 27     & Handbag           & 28     & Tie               \\
29      & Suitcase          & 30     & Frisbee           & 31     & Skis              & 32     & Snowboard         \\
33      & Sports ball       & 34     & Kite              & 35     & Baseball bat      & 36     & Baseball glove    \\
37      & Skateboard        & 38     & Surfboard         & 39     & Tennis rocket     & 40     & Bottle            \\
41      & Wine glass        & 42     & Cup               & 43     & Fork              & 44     & Knife             \\
45      & Spoon             & 46     & Bowl              & 47     & Banana            & 48     & Apple             \\
49      & Sandwich          & 50     & Orange            & 51     & Broccoli          & 52     & Carrot            \\
53      & Hot dog           & 54     & Pizza             & 55     & Donut             & 56     & Cake              \\
57      & Chair             & 58     & Couch             & 59     & Potted plant      & 60     & Bed               \\
61      & Dining table      & 62     & Toilet            & 63     & TV                & 64     & Laptop            \\
65      & Mouse             & 66     & Remote            & 67     & Keyboard          & 68     & Cell phone        \\
69      & Microwave         & 70     & Oven              & 71     & Toaster           & 72     & Sink              \\
73      & Refrigerator      & 74     & Book              & 75     & Clock             & 76     & Vase              \\
77      & Scissors          & 78     & Teddy bear        & 79     & Hair drier        & 80     & Toothbrush       
\end{tabular}
\caption{Category splits ($S_1$ -- $S_4$, Section~\ref{sec:task}) of \coco.}
\label{table:dataset_splits}
\end{table}

\subsection{Rationale}

Providing four splits with equally distributed held-out categories has two main advantages: It allows to test on all categories in \coco (albeit with different models) while sub sampling the super categories~\cite{Lin2014} as evenly as possible. This approach assumes that we will know some objects from all broad object categories in the world and that we can infer the missing parts from this knowledge. This setup differs from tasks like \emph{tiered}ImageNet~\cite{Ren2018} which require generalization to objects from vastly different categories.

\section{Implementation details}
\label{appendix:implementation}

\subsection{Backbone}
We use the standard architecture of ResNet-50 \cite{He2016} without any modifications.

\subsection{Feature matching}
\begin{itemize}[nosep]
    \item We use layers\footnote{Using the notation from here: \url{https://ethereon.github.io/netscope/##/gist/db945b393d40bfa26006}} \texttt{res2c\_relu} (256 features), \texttt{res3d\_relu} (512), \texttt{res4f\_relu} (1024) and \texttt{res5c\_relu} (2048) of the backbone as a feature representation of the inputs. For brevity, we refer to these layers as $C_2$, $C_3$, $C_4$ and $C_5$.
    \item FPN generates multi-scale representations $P_i$, $i = \{2,3,4,5, 6\}$ consisting of 256 features (for all $i$) as follows. $P_5$ is a result of applying a $1 \times 1$ conv layer to $C_5$ (to get 256 features). $P_i$ ($i = \{2,3,4\}$) is a sum of a $1\times1$ conv layer applied to $C_i$ and up-sampled (by a factor of two on each side) $P_{i+1}$. $P_6$ is a down-sampled $P_5$ (by a factor of two on each side).
    \item The final similarity scores between the input scene and the reference at scale $i$ are computed by obtaining $P_i^{\,\text{scene}}$ and $P_i^{\,\text{ref}}$ as described above, applying global average pooling to $P_i^{\text{ref}}$, and computing pixel-wise differences $D_i = \text{abs}(P_i^{\,\text{scene}} - \text{pool}(P_i^{\text{ref}}))$. 
    \item The final feature representations containing information about similarities between the scene and the reference are computed by concatenating $P_i^{\,\text{scene}}$ and $D_i$, and applying a $1\times1$ conv layer, outputting 384 features.
\end{itemize}

\subsection{Region Proposal Network (RPN)}
\begin{itemize}[nosep]
    \item We use 3 anchor aspect ratios (0.5, 1, 2) at each pixel location for the 5 scales (32, 64, 128, 256, 512) $i = \{2, \ldots, 6\}$ defined above, resulting in $3 \times (32^2 + \ldots + 512^2) \approx 1\text{M}$ proposals in total.
    \item The architecture is a $3 \times 3 \times 512$ conv layer, followed by the $1 \times 1$ conv outputting $k$ times number of anchors per location (three in our case) features (corresponding to proposal logits for $k = 2$ or to bounding box deltas for $k = 4$).
\end{itemize}

\subsection{Classification and bounding box regression head}
The classification head produces same/different classifications for each proposal and performs bounding box regression.

\begin{itemize}[nosep]
    \item Inputs: the computed bounding boxes (outputs of the RPN) are cropped from  $P_i$, reshaped to $7 \times 7$, and concatenated for $i = \{2,\ldots,5\}$. Only 6000 top scoring anchors are processed for efficiency.
    \item Architecture: two fc-layers (1024 units with ReLU) followed by a logistic regression into 2 classes (same as reference or not).
    \item Bounding box regression is part of the classification branch, but uses a different output layer. This output layer produces fine adjustments (deltas) of the bounding box coordinates (instead of class probabilities).
    \item Non-maximum suppression (NMS; threshold 0.7) is applied to the predicted bounding boxes.
\end{itemize}

\subsection{Segmentation head}
\begin{itemize}[nosep]
    \item Inputs: the computed bounding boxes are cropped from  $P_i$, reshaped to $14 \times 14$, and concatenated for $i = \{2,\ldots,5\}$.
    \item Architecture: four $3\times3$ conv layers (with ReLU and BN) followed by a transposed conv layer with $2\times2$ kernels and stride of 2, and a final $1\times1$ conv layer outputting two feature maps consisting of logits for foreground/background at each spatial location.
\end{itemize}

\section{Training details}
\label{appendix:training}

This section contains a detailed description of the training procedure. To make this section more readable and have all relevant information in one place it contains a few duplications with Section~\ref{subsec:training}

\paragraph{Pre-training backbone.}
We pre-train the ResNet backbone on image classification on a reduced subset of ImageNet, which contains images from the 687 ImageNet categories without correspondence in \coco~-- hence we refer to it as \emph{ImageNet-687}. Pre-training on this reduced set ensures that we do not use any label information about the test categories at any training stage.

\paragraph{Training Siamese Mask R-CNN.}
We train the models using stochastic gradient descent with momentum for 160,000 steps with a batch size of 12 on 4 NVIDIA P100 GPUs in parallel. With this setup training takes roughly a week. We use an initial learning rate of 0.02 and a momentum of 0.9. We start our training with a warm-up phase of 1,000 steps during which we train only the heads. After that, we train the entire network, including the backbone and all heads, end-to-end. After 120,000 steps, we divide the learning rate by 10.

\paragraph{Construction of mini-batches.}
During training, a mini-batch contains 12 sets of reference and query images. We first draw the query images at random from the training set and pre-process them in the following way: (1) we resize an image so that the longer side is 1024~px, while keeping the aspect ratio, (2) we zero-pad the smaller side of the image to be square $1024 \times 1024$, (3) we subtract the mean ImageNet RGB value from each pixel. Next, for each image, we generate a reference image as follows: (1) draw a random category among all categories of the background set present in the image, (2) crop a random instance of the selected category out of any image in the training set (using the bounding box annotation), and (3) resize the reference image so that its longer side is 192~px and zero-pad the shorter side to get a square image of $192 \times 192$. To enable a quick look-up of reference instances, we created an index that contains a list of categories present in each image.

\paragraph{Labels.}
We use only the annotations of object instances in the query image that belong to the corresponding reference category. The annotations of all other objects are removed and subsequently they are treated as background.

\paragraph{Loss function.}

Siamese Mask R-CNN is trained on the same basic multi-task objective as Mask R-CNN: classification and bounding box loss for the RPN; classification, bounding box and mask loss for each RoI. There are a couple of differences as well. First, the classification losses consist of a binary cross-entropy of the match/non-match classification rather than an 80-way multinomial cross-entropy used for classification on \coco. Second, we found that weighting the individual losses differently improved performance in the one-shot setting. Specifically, we apply the following weights to each component of the loss function: RPN classification loss: 2, RPN bounding box loss: 0.1, RoI classification loss: 2, RoI bounding box loss: 0.5 and mask loss: 1.


\paragraph{Exact hyper parameter details}
Complex systems like Mask R-CNN require a large set of hyper parameters to be set for optimal training performance. We mentioned all changes we made to the hyperparameter settings of the implementation we extended~\cite{Abdulla2017}. For the full list of hyperparameter settings and exact details of our loss function implementation and data handling please refer to the code: \url{https://github.com/bethgelab/siamese-mask-rcnn}

\section{Evaluation details}
\label{appendix:evaluation}

This section contains a detailed description and discussion of the evaluation procedure. As with the training section it contains a few duplications with the corresponding Section~\ref{sec:task} from the main paper in order to have all information in one place.

\subsection{Category selection}
The evaluation is performed on the MS Coco 2017 validation set (which corresponds to the 2014 minval set). The evaluation is performed for 4 subtasks, each using 60 categories for training and the remaining 20 categories for one-shot evaluation. Those 20 categories are selected by choosing every 4th category, therefore the $i$th split is constructed by: $[i + 4*k \mathrm{~for~} k \mathrm{~in~range~}(20)]$. An explicit listing of all 4 splits can be found in Table~\ref{table:dataset_splits} above.

\subsection{Evaluation procedure}
Each of the subtasks is evaluated over the whole validation set using the corresponding set of categories. Therefore for each image the present categories from the current split are determined. Then for each present category a reference instance is randomly chosen from the whole evaluation set (those references are chosen individually for each image). The model is then evaluated for each of the references and the predictions of each of these runs is assigned to the corresponding category. If no category from the current split is present the image is skipped. After running this over all images the results contain predicted bounding boxes for each image but only for the categories of the selected split. These collected results can then be fed to a slightly modified version of the official \coco analysis tools \cite{Coco2018} which can handle specific category subsets to get the final mAP50 scores.

\begin{algorithm}[H]
 \For{image \textbf{in} images}{
  present categories = get one shot categories(image)\;
  \For{category \textbf{in} present categories}{
   ref = get random instance (category, images)\;
   results[image, category] = model.predict (ref, image)\;
  }
 }
 mAP50 = evaluate mAP50(results, one shot categories)\;
 \vspace{6pt}
 \caption{Pseudocode for evaluation procedure}
\end{algorithm}

\subsection{Noise induced by random reference sampling}
Because only one reference is sampled per category and image the predictions can be rather noisy (especially in the one-shot case). For our model the std of the predicted results is $\pm1\%$. To get a good prediction of the actual mean we run the evaluation of each split 5 times thus reaching reaching a standard error of the mean of less than $\pm0.2\%$.

\subsection{Comment on the evaluation procedure}
We specifically chose to evaluate our model only on the categories present in each image. We think, that this scenario can realistically be assumed in real world tasks as a whole-image classification network can be used to pre-select if the reference category is present in an image before running the bounding box and instance segmentation prediction network.

This choice, however, makes the task substantially easier than evaluating each image for all categories. It does not punish false positives as hard as the other task does. However, as visible in our results, false positives play an important role even in our simpler task, which leads us to the conclusion, that our task setup is still sufficiently difficult.

\subsection{Note on non-maximum suppression}
We use non-maximum suppression (NMS) on the predictions of each image/references combination individually and not on the combined output of an image after running the detection for all references because at test time the system needs to be able to detect and segment objects based on only a single reference example of each category separately.

\subsection{Choice of evaluation metric}

We chose to use mAP50 instead of the so called "coco metric" mAP. mAP50 is evaluated at a single Intersection over Union (IoU) threshold of 50\% between predicted and the ground truth bounding boxes (corresponding to around 70\% overlap between two same-sized boxes/masks) while mAP is evaluated at IoU thresholds of [50\%, 55\%, ..., 95\%] adding weight to exact bounding box/segmentation mask predictions.

We think, that mAP50 is the value most reflective of the result we are interested in: whether our model can find novel objects based on a single reference image. For instance segmentation the additional information about mask quality implicitly included in mAP might make sense. However we found, that correctly masking the sought objects was less of a problem for our model than correctly classifying them.

\section{Confusion matrix}
\label{appendix:confusion_matrix}

\begin{figure}[t!]
    \centering
    \includegraphics[width=\linewidth]{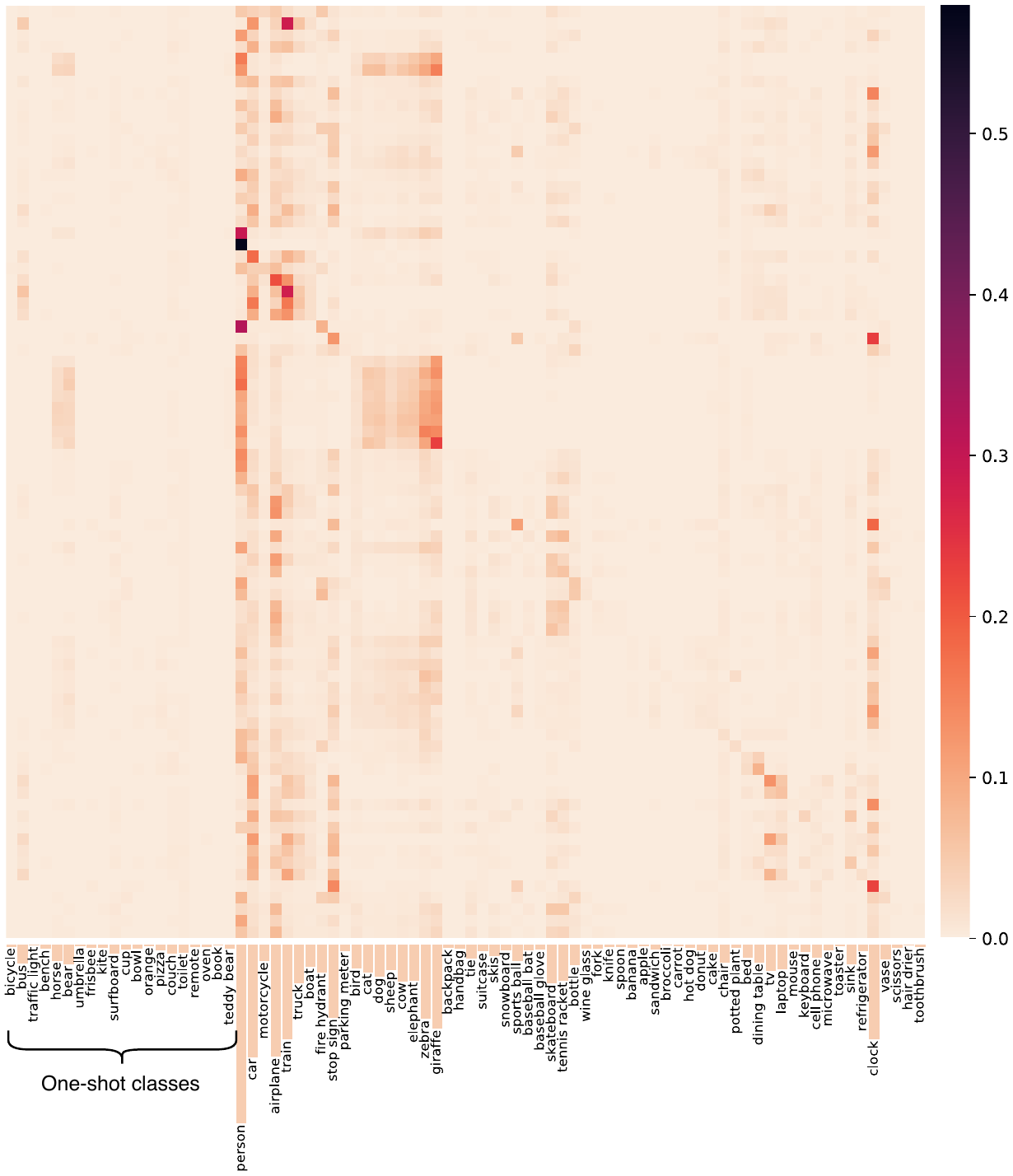}
    \caption{Confusion matrix for the Siamese Mask R-CNN model using split $S_2$ for one-shot evaluation. The element $(i,j)$ shows the AP50 of using detections for category $i$ and evaluating them as instances of category $j$. The histogram below the matrix shows the most commonly confused (or falsely predicted) categories.}
    \label{fig:confusion_matrix}
\end{figure}

To quantify the errors of our model we compute a confusion matrix over the 80 categories in \coco using a model trained on split S$^2$ (Figure~\ref{fig:confusion_matrix}). The element $(i,j)$ of this matrix corresponds to the AP50 value of detections obtained for reference images of category $i$, which are evaluated as if the reference images belonged to category $j$. If there were no false positives, the off-diagonal elements of the matrix would be zero. The sums of values in the columns show instances of categories that are most often falsely detected (the histogram of such sums is shown below the matrix). Among such commonly falsely predicted categories are people, cars, airplanes, clocks, and other categories that are common in the dataset.

\section{Additional results}
\label{appendix:additional_results}

In this section we discuss the noisiness of our evaluation approach and provide additional results including split-by-split values for the 95\% confidence intervals we get from running the evaluation 5 times (Table~\ref{table:results_by_split}) and the full results on all metrics evaluated on the \coco leaderboard (\url{cocodataset.org/\#detection-leaderboard}) for object detection (Tables~\ref{table:extended_one-shot_detection_results}~\&~\ref{table:extended_five-shot_detection_results}) and instance segmentation (Tables~\ref{table:extended_one-shot_segmentation_results}~\&~\ref{table:extended_five-shot_segmentation_results}).

\subsection{Noisiness of evaluation}

The example based evaluation setting with a randomly drawn reference per category and image is naturally prone to be noisy. We therefore evaluate our models five times and take the mean of these 5 evaluations as our final result. We here want to discuss the amount of randomness generated by our evaluation procedure and the confidence of our mean.

We found the standard deviation of one-shot object detection and instance segmentation segmentation to be around 0.3\% mAP50 while the standard deviation with five reference images is lower at 0.1\% mAP50. The 95\% confidence of the mean is around 0.1\% (See Table~\ref{table:results_one_shot}. The rather small deviations can be seen as a result of the evaluation procedure which considers every image and reference category as a single instance. This ensures that there are many samples per category over test set.

\subsection{Results for each split}

We show the results for each split (S$^1$-S$^4$) separately reporting mean and 95\% confidence interval of five evaluation runs in Table~\ref{table:results_by_split}. We find slight difference in performance between these split with split S$^1$ showing the biggest gap between evaluating on the training and test categories. We assume, that this is due to the strong over representation of the person category in \coco \cite{Lin2014}. With a lot of small instances and presence of persons in almost every image the removal of this category during training makes the dataset considerably easier, while requesting to detect them later is hard.

\begin{table}[h]
\begin{center}
\begin{small}
\begin{tabular}{lcccccc}
 \multicolumn{6}{c}{Object detection\vspace{4pt}}\\
Categories & Shots & S$^1$ & S$^2$ & S$^3$ & S$^4$ & \O\\
\hline
\multirow{2}{*}{Train} & 1 & $39.1\plusminus{0.1}$ & $36.6\plusminus{0.1}$ & $37.5\plusminus{0.1}$ & $37.2\plusminus{0.2}$ & $37.6\plusminus{0.1}$ \\
 & 5 & $42.4\plusminus{0.1}$ & $40.5\plusminus{0.1}$ & $41.5\plusminus{0.1}$ & $40.9\plusminus{0.2}$ & $41.3\plusminus{0.1}$ \\
\hline
\multirow{2}{*}{Test} & 1 & $15.3\plusminus{0.2}$ & $16.8\plusminus{0.2}$ & $16.7\plusminus{0.2}$ & $16.4\plusminus{0.1}$ & $16.3\plusminus{0.1}$ \\ 
 & 5 & $16.8\plusminus{0.1}$ & $20.0\plusminus{0.1}$ & $18.2\plusminus{0.1}$ & $19.0\plusminus{0.1}$ & $18.5\plusminus{0.1}$ \\
\multicolumn{6}{c}{}\\
 \multicolumn{6}{c}{Instance segmentation\vspace{4pt}} \\
Categories  & Shots & S$^1$ & S$^2$ & S$^3$ & S$^4$ & \O\\
 \hline
\multirow{2}{*}{Train} & 1 & $36.6\plusminus{0.1}$ & $33.5\plusminus{0.1}$ & $34.9\plusminus{0.1}$ & $34.5\plusminus{0.2}$ & $34.9\plusminus{0.1}$ \\
 & 5 & $39.7\plusminus{0.1}$ & $37.3\plusminus{0.1}$ & $38.7\plusminus{0.1}$ & $37.9\plusminus{0.2}$ & $38.4\plusminus{0.1}$ \\
\hline
\multirow{2}{*}{Test} & 1 & $13.5\plusminus{0.2}$ & $14.9\plusminus{0.1}$ & $15.5\plusminus{0.2}$ & $14.2\plusminus{0.1}$ & $14.5\plusminus{0.1}$ \\
 & 5 & $14.8\plusminus{0.1}$ & $18.0\plusminus{0.1}$ & $17.4\plusminus{0.1}$ & $16.9\plusminus{0.1}$ & $16.8\plusminus{0.1}$ \\
\end{tabular}
\end{small}
\end{center}
\caption{Results on MS Coco (in \% mAP50 with 95\% confidence intervals). In split S$^i$, every fourth category, starting at the $i^\mathrm{th}$, is placed into the test set.}
\label{table:results_by_split}
\end{table}

\subsection{Full \coco style results}

In this section we report results on all metrics used at the \coco leader board \url{cocodataset.org/\#detection-leaderboard}. Beyond the mAP50 (AP$^{50}$) metric reported in the main paper these include the \coco metric (AP) as well as other AP metrics at different thresholds (AP$^{75}$) and object sizes (AP$^\text{S}$, AP$^\text{M}$, AP$^\text{L}$ each as subsets of AP) as well as recall metrics (AR) with varying numbers of detections (AR$^{1}$, AR$^{10}$, AR$^{100}$) and object sizes (AR$^\text{S}$, AR$^\text{M}$, AR$^\text{L}$ each as parts of AR$^{100}$).

\begin{table}[h]
\begin{center}
\begin{small}
\begin{tabular}{c|ccc|ccc|ccc|ccc}
Model & AP & AP$^{50}$ & AP$^{75}$ & AP$^\text{S}$ & AP$^\text{M}$ & AP$^\text{L}$ & AR$^{1}$ & AR$^{10}$ & AR$^{100}$ & AR$^\text{S}$ & AR$^\text{M}$ & AR$^\text{L}$\\
\hline
full & 21.8 & 35.5 & 23.4 & 11.1 & 21.8 & 30.8 & 19.9 & 37.6 & 39.2 & 22.2 & 41.0 & 56.5 \\
\hline
train S$^1$ & 23.6 & 39.1 & 25.0 & 11.4 & 23.3 & 33.8 & 20.9 & 38.9 & 40.7 & 22.9 & 43.1 & 57.5 \\
train S$^2$ & 21.9 & 36.6 & 23.5 & 11.4 & 22.6 & 31.1 & 19.9 & 37.9 & 39.4 & 22.7 & 41.9 & 57.1 \\
train S$^3$ & 23.3 & 37.5 & 25.2 & 11.1 & 22.5 & 33.4 & 20.9 & 39.3 & 41.0 & 21.8 & 43.1 & 59.7 \\
train S$^4$ & 22.7 & 37.2 & 24.2 & 11.9 & 21.6 & 31.7 & 20.1 & 38.5 & 40.4 & 23.2 & 42.4 & 56.7 \\
\hline
test S$^1$ & 8.6 & 15.3 & 8.8 & 5.0 & 8.6 & 13.5 & 10.3 & 26.4 & 27.7 & 14.4 & 29.9 & 43.2 \\
test S$^2$ & 9.8 & 16.8 & 10.1 & 5.7 & 8.4 & 14.8 & 12.2 & 26.7 & 27.7 & 13.9 & 27.6 & 43.9 \\
test S$^3$ & 8.9 & 16.7 & 8.8 & 5.6 & 8.2 & 16.6 & 9.4 & 23.6 & 24.6 & 15.3 & 25.1 & 40.0 \\
test S$^4$ & 9.1 & 16.4 & 9.2 & 5.4 & 9.4 & 14.0 & 10.9 & 25.7 & 27.4 & 14.5 & 30.7 & 43.2
\end{tabular}
\end{small}
\end{center}
\caption{Full \textbf{one-shot detection} results on \coco. train/test indicate evaluation on the training/test categories of split S$^i$ respectively. Each value is the mean of 5 evaluation runs.}
\vspace{-12pt}
\label{table:extended_one-shot_detection_results}
\end{table}

\begin{table}[h]
\begin{center}
\begin{small}
\begin{tabular}{c|ccc|ccc|ccc|ccc}
Model & AP & AP$^{50}$ & AP$^{75}$ & AP$^\text{S}$ & AP$^\text{M}$ & AP$^\text{L}$ & AR$^{1}$ & AR$^{10}$ & AR$^{100}$ & AR$^\text{S}$ & AR$^\text{M}$ & AR$^\text{L}$\\
\hline
full & 19.3 & 33.1 & 19.9 & 9.3 & 19.4 & 27.5 & 17.9 & 33.5 & 34.9 & 19.4 & 36.9 & 49.9 \\
\hline
train S$^1$ & 20.9 & 36.6 & 21.0 & 9.3 & 20.9 & 30.5 & 19.0 & 34.6 & 36.1 & 19.6 & 38.9 & 51.3 \\
train S$^2$ & 18.9 & 33.5 & 19.4 & 9.2 & 19.3 & 27.4 & 17.8 & 33.2 & 34.5 & 19.6 & 36.7 & 50.0 \\
train S$^3$ & 20.0 & 34.9 & 20.2 & 9.0 & 19.5 & 29.6 & 18.7 & 34.8 & 36.2 & 18.6 & 38.4 & 53.9 \\
train S$^4$ & 20.0 & 34.5 & 20.9 & 9.9 & 19.0 & 28.6 & 18.2 & 34.3 & 35.7 & 20.3 & 37.5 & 51.2 \\
\hline
test S$^1$ & 6.7 & 13.5 & 6.0 & 3.8 & 6.8 & 11.0 & 8.3 & 22.0 & 23.0 & 11.7 & 25.2 & 36.1 \\
test S$^2$ & 8.5 & 14.9 & 8.7 & 4.7 & 7.4 & 12.8 & 10.8 & 23.5 & 24.5 & 11.7 & 24.7 & 39.3 \\
test S$^3$ & 8.2 & 15.5 & 8.0 & 4.7 & 7.2 & 15.3 & 9.0 & 21.8 & 22.7 & 13.9 & 22.8 & 35.9 \\
test S$^4$ & 7.3 & 14.2 & 6.6 & 3.8 & 7.9 & 11.8 & 9.3 & 22.3 & 23.8 & 12.0 & 28.2 & 38.2
\end{tabular}
\end{small}
\end{center}
\caption{Full \textbf{one-shot segmentation} results on \coco. train/test indicate evaluation on the training/test categories of split S$^i$ respectively. Each value is the mean of 5 evaluation runs.}
\vspace{-12pt}
\label{table:extended_one-shot_segmentation_results}
\end{table}

\begin{table}[h]
\begin{center}
\begin{small}
\begin{tabular}{c|ccc|ccc|ccc|ccc}
Model & AP & AP$^{50}$ & AP$^{75}$ & AP$^\text{S}$ & AP$^\text{M}$ & AP$^\text{L}$ & AR$^{1}$ & AR$^{10}$ & AR$^{100}$ & AR$^\text{S}$ & AR$^\text{M}$ & AR$^\text{L}$\\
\hline
full & 24.9 & 40.5 & 26.7 & 13.3 & 25.0 & 35.9 & 21.8 & 40.1 & 41.8 & 23.9 & 44.3 & 59.1 \\
\hline
train S$^1$ & 25.7 & 42.4 & 27.1 & 12.6 & 25.6 & 36.2 & 22.1 & 40.6 & 42.4 & 24.3 & 45.1 & 59.3 \\
train S$^2$ & 24.3 & 40.5 & 26.1 & 12.8 & 25.1 & 35.3 & 21.4 & 39.7 & 41.3 & 24.1 & 44.2 & 59.9 \\
train S$^3$ & 25.8 & 41.5 & 28.0 & 12.7 & 25.2 & 38.2 & 22.4 & 41.0 & 42.7 & 23.5 & 45.1 & 61.5 \\
train S$^4$ & 25.1 & 40.9 & 26.8 & 12.9 & 23.8 & 36.3 & 21.5 & 40.3 & 42.3 & 24.7 & 44.5 & 59.1 \\
\hline
test S$^1$ & 9.4 & 16.8 & 9.7 & 5.6 & 9.3 & 14.6 & 11.0 & 28.1 & 29.4 & 15.8 & 31.9 & 45.8 \\
test S$^2$ & 11.7 & 20.0 & 12.1 & 6.3 & 9.7 & 19.3 & 13.3 & 29.1 & 30.3 & 15.1 & 30.7 & 48.3 \\
test S$^3$ & 9.8 & 18.2 & 9.5 & 6.7 & 9.2 & 17.5 & 9.6 & 25.0 & 26.0 & 16.3 & 26.4 & 42.4 \\
test S$^4$ & 10.6 & 19.0 & 10.6 & 5.8 & 10.4 & 16.6 & 11.8 & 27.8 & 29.6 & 14.8 & 33.1 & 47.5
\end{tabular}
\end{small}
\end{center}
\caption{Full \textbf{five-shot detection} results on \coco. train/test indicate evaluation on the training/test categories of split S$^i$ respectively. Each value is the mean of 5 evaluation runs.}
\vspace{-12pt}
\label{table:extended_five-shot_detection_results}
\end{table}

\begin{table}[h]
\begin{center}
\begin{small}
\begin{tabular}{c|ccc|ccc|ccc|ccc}
Model & AP & AP$^{50}$ & AP$^{75}$ & AP$^\text{S}$ & AP$^\text{M}$ & AP$^\text{L}$ & AR$^{1}$ & AR$^{10}$ & AR$^{100}$ & AR$^\text{S}$ & AR$^\text{M}$ & AR$^\text{L}$\\
\hline
full & 22.0 & 37.8 & 22.9 & 11.4 & 22.2 & 32.4 & 19.7 & 35.8 & 37.2 & 21.0 & 39.5 & 52.6
 \\
\hline
train S$^1$ & 22.7 & 39.7 & 23.1 & 10.3 & 23.0 & 33.0 & 20.2 & 36.1 & 37.6 & 20.8 & 40.7 & 53.2 \\
train S$^2$ & 21.0 & 37.3 & 21.7 & 10.5 & 21.7 & 31.2 & 19.2 & 34.9 & 36.3 & 20.7 & 38.7 & 52.9 \\
train S$^3$ & 22.4 & 38.7 & 22.8 & 10.3 & 21.9 & 34.2 & 20.3 & 36.5 & 37.8 & 19.8 & 40.2 & 56.3 \\
train S$^4$ & 22.1 & 37.9 & 23.2 & 10.6 & 20.8 & 32.9 & 19.5 & 35.9 & 37.6 & 21.4 & 39.3 & 53.6
 \\
\hline
test S$^1$ & 7.4 & 14.8 & 6.7 & 4.3 & 7.2 & 12.2 & 9.2 & 23.7 & 24.7 & 13.1 & 26.8 & 39.3 \\
test S$^2$ & 10.2 & 18.0 & 10.5 & 5.1 & 8.6 & 17.2 & 12.0 & 26.0 & 27.0 & 12.5 & 27.7 & 44.1 \\
test S$^3$ & 9.0 & 17.4 & 8.5 & 5.6 & 8.2 & 16.6 & 9.4 & 23.0 & 23.9 & 14.5 & 24.3 & 38.3 \\
test S$^4$ & 8.5 & 16.9 & 7.8 & 4.1 & 8.8 & 14.4 & 10.3 & 24.3 & 25.9 & 12.3 & 30.4 & 42.0
\end{tabular}
\end{small}
\end{center}
\caption{Full \textbf{five-shot segmentation} results on \coco. train/test indicate evaluation on the training/test categories of split S$^i$ respectively. Each value is the mean of 5 evaluation runs.}
\vspace{-12pt}
\label{table:extended_five-shot_segmentation_results}
\end{table}

\end{document}